\RequirePackage{fix-cm}
\documentclass[twocolumn]{svjour3}          % twocolumn
\smartqed  % flush right qed marks, e.g. at end of proof
\usepackage{graphicx}
\usepackage{mathptmx}      % use Times fonts if available on your TeX system

\usepackage{hyperref}
\usepackage{xcolor}
\usepackage{amssymb,amsmath}

\begin{document}

    \title{Unsuitability of NOTEARS for Causal Graph Discovery}

    \author{Marcus Kaiser \and Maksim Sipos}

    \institute{M. Kaiser \at
    causaLens \\
    \email{marcus@causalens.com}
    \and
    M. Sipos \at
    causaLens \\
    \email{max@causalens.com}
    }

    \date{Received: date / Accepted: date}

    \maketitle

    \begin{abstract}

        Causal Discovery methods aim to identify a DAG structure that represents causal relationships from observational data.
        In this article, we stress that it is important to test such methods for robustness in practical settings.
        As our main example, we analyze the NOTEARS method, for which we demonstrate a lack of scale-invariance.
        We show that NOTEARS is a method that aims to identify a parsimonious DAG from the data that explains the residual variance.
        We conclude that NOTEARS is not suitable for identifying truly causal relationships from the data.

    \end{abstract}

    \section{Introduction}

    Establishing the relationship between cause and effect normally requires a randomized controlled trial (RCT),
    or an ability to perform an intervention~\cite{peters2017elements}.
    However, this is not always possible or ethical in practical settings.
    The field of Causal Discovery aims to identify causal relationships from purely
    observational data. For tabular data with independently and identically distributed (IID) samples, causal
    relationships can conveniently be represented by a Directed Acyclic Graph
    (DAG)~\cite{glymour2019review,peters2017elements}.
    Thus the aim of many Causal Discovery methods is to generate a
    DAG that resembles the true causal relationships in the data.
    For most real world use cases, there typically is no ground truth one
    can compare results to and it is only partially possible to verify the
    identified cause and effect relationships. Therefore, it is important
    that methods deployed on such data are robust with respect to
    changes in hyperparameters~\cite{dgp_causal_discovery_time_series}, and invariant with respect to
    reparameterization of the data.
    When it comes to applicability of newly developed methods to real data, there are the following pitfalls:

    \begin{itemize}
        \item
        Evaluation on examples for which the method has been
        developed (and is expected to perform well). It is important to also
        test how the method behaves when some of the assumptions (such as
        linearity, additivity, homoscedasticity and causal sufficiency~\cite{peters2017elements}) are not satisfied.
        \item
        No checks for robustness to changes in hyperparameters.
        In order to be able to trust the results,
        it is important that the results either are only mildly affected by changes of hyperparameters,
        or there are reasonable heuristics for how to choose these hyperparameters.
        \item
        No checks for invariance with respect to the parameterization of the
        data. For example, how does the method behave when the scale of some
        of the variables is altered (such as a change of units, for example from meters to centimeters).
    \end{itemize}

    For the rest of the paper, we focus on the third point.
    We analyze the recent NOTEARS~\cite{zheng2018dags} method and show how a
    simple rescaling of the variables can lead to a large alteration of the derived graph.

    \section{The NOTEARS method}

    One popular method (with 100+ citations) for extracting a DAG structure from observational data published recently is NOTEARS
    (with open source code available from \url{https://github.com/xunzheng/notears}).
    Given a loss function $\ell(W; X)$, the core idea of this method is to optimize the objective function
    \begin{equation}\label{eqn:full_f}
        F(W) := \ell(W; X) + \lambda \|W\|_1,
    \end{equation}
    with the additional constraint that the weight matrix $W$ has to represent a DAG. $\| \cdot \|_1$ is the $L_1$ matrix norm
    and $\lambda \ge 0$ is a regularization parameter that controls the sparsity of the identified weights.
    Following the suggestion in~\cite{zheng2018dags}, we concentrate on the least-squares loss, given by
    \begin{equation}\label{eqn:LS}
        \ell(W; X) := \frac 1{2n} \|X - X W \|_F^2,
    \end{equation}
    where $\| \cdot \|_F$ is the Frobenius norm or $L_2$ matrix norm and $n$ is the total number of samples. We come back to the choice of $\ell(W; X)$ in Section \ref{sec:relatedwork} below.
    Theorem 1 in~\cite{zheng2018dags} states that $W \in \mathbb R^{d \times d}$ is a DAG if and only if
    \begin{equation}\label{eqn:h}
        h(W) := \operatorname{trace}\Bigl (\mathrm e^{W \circ W}\Bigr) - d = 0,
    \end{equation}
    where $A\circ B$ is the Hadamard product (entry-wise product of matrix entries).
    Based on this insight, the authors of NOTEARS define the following constrained non-linear problem
    \begin{equation}
        \label{eqn:opt_problem}
        \min_{W \in \mathbb R^{d \times d}} F(W) \quad \textrm{subject to} \quad h(W) = 0.
    \end{equation}
    Problems of the form \eqref{eqn:opt_problem} can be solved using an Augmented
    Lagrangian~\cite{fortin2000augmented} approach,
    which replaces the constrained problem with a
    sequence of unconstrained problems. It is given by
    \begin{equation}
        \label{eqn:opt_aug_lagrangian}
        \min_{W \in \mathbb R^{d \times d}} F(W) + \frac \rho2 |h(W)|^2 + \alpha h(W),
    \end{equation}
    where $\alpha > 0$ and $\rho > 0$ are iteratively increased, in order to find solutions that minimize $h(W)$ (see~\cite{zheng2018dags} for the full details).
    Asymptotically, this leads to an (approximate) DAG structure for $W$.
    The whole algorithm can then be seen as an outer update step, where $\alpha$ and $\rho$ are
    increased, as well as an inner update step, which can e.g.\ be solved with a second order Newton method, such as L-BFGS-B~\cite{zhu1997algorithm}.
    There are a few points worth noting here:
    \begin{itemize}
        \item
        The overall algorithm is non-convex, which implies that the method
        can have multiple minima (and potentially can get stuck in such a
        local minimum).
        \item
        The method tries to find a trade-off between three objectives, which
        are an optimal fit of the linear model $X = XW$, whilst $W$
        should both resemble a DAG and be sparse.\\
        Note that the first objective is opposed to the second and third
        objective, which both are aligned to one another as it is easier for a
        graph to not violate the DAG constraints when it has fewer edges that
        could violate these assumptions.
        \item
        The (approximate) DAG structure returned by the algorithm does not
        necessarily represent any causal relationships between the variables,
        but rather is the estimated result of a parsimonious model of the form of a linear Structural Equation Model (SEM)~\cite{glymour2019review}.
    \end{itemize}

    Although the authors in~\cite{zheng2018dags} do not claim that the estimated SEM represents causal relationships, one can find many
    references in the literature, which equate the structure learning approach from NOTEARS with learning causal graphs or Bayesian networks,
    cf.~\cite{ng2019masked,yu2019dag,zheng2020learning,pamfil2020dynotears,bhattacharya2021differentiable,kyono2020castle}.    
    This however is not appropriate because identifying parsimonious structures that explain the data is not equivalent to making statements about
    causality in the data generating process, cf.\ the discussion in~\cite[Sec.~6.2]{vowels2021d}.
    Causal graphs imply more information than that which is present within a statistical model such as a Bayesian Network, 
    most notably the ability to model interventions upon the data using do-calculus~\cite{pearl2012calculus,scholkopf2019causality}.

    \section{Illustrative example}

    We consider a simple example consisting of a linear SEM with four variables
    $X = (X_0, X_1,X_2, X_3)$, uniform noise and $n=1000$.
    Each variable is centered to be mean zero.
    This setup satisfies the assumptions of NOTEARS and therefore the method is
    expected to perform well. We want to understand how this
    method is affected by rescaling the data. We stress that the below
    described issues do not depend on this particular setup, but are general.

    \begin{figure}
        \includegraphics[scale=0.27]{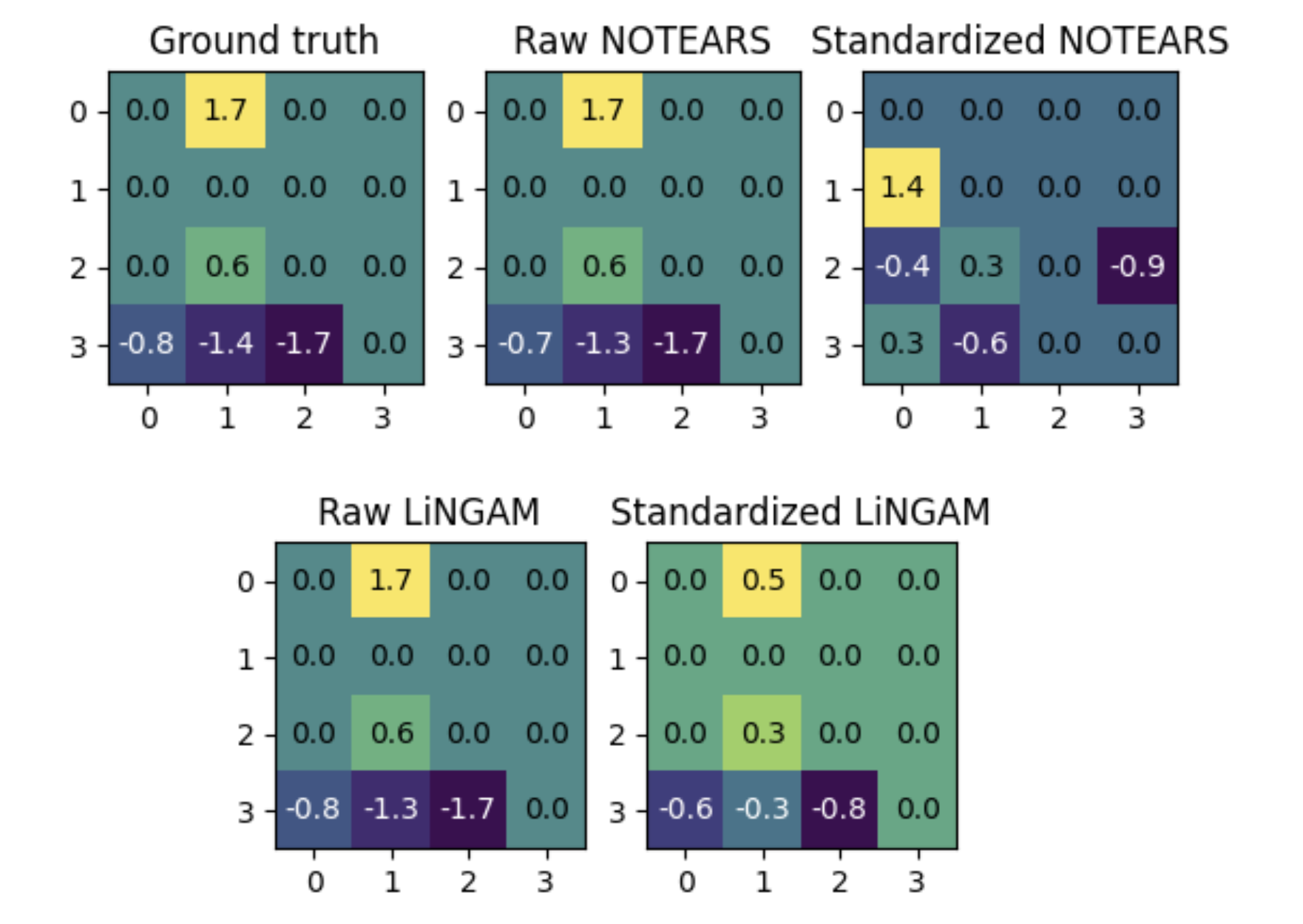}
        \caption{
            The weights of the simulated process. The entries in
            the $i$-th column represent how much variable $i$ is affected by all other variables. Top row: We show the
            ground truth, results for the original generated data, as well as results for standardized data.
            Bottom row: Results for the LiNGAM algorithm (for comparison).}
        \label{fig:weights_comparison}
    \end{figure}

    Figure~\ref{fig:weights_comparison} presents five plots. The first row shows the underlying ground truth $W$ used to generate the data,
    followed by two iterations of NOTEARS. It is applied to the raw data (with standard deviations $\sigma(X) = (0.86, 1.56, 1.07, 0.76)$),
    as well as normalized (unit standard deviation for all variables).
    The bottom row shows the corresponding results for the LiNGAM~\cite{shimizu2011directlingam} algorithm (based on the DirectLiNGAM
    implementation available at \url{https://github.com/cdt15/lingam}).
    We can observe that the result from NOTEARS on the original scale (second plot on the top) is close to the ground truth,
    but the identified variables for normalized data are not correct (top right).
    One can see that the weights for $X_0$ and $X_1$ are ``flipped'', resulting in the method identifying that $X_1 \rightarrow X_0$, as opposed
    to $X_0 \rightarrow X_1$ (cf. the description of Figure~\ref{fig:weights_comparison} for how to interpret the coefficients).
    In comparison, LiNGAM is able to correctly identify the variables.
    (For the scaled variables, the resulting weights differ due to the scaling, but the identified DAGs coincide.)

    Note that for $\lambda > 0$, one would expect the scaling to affect the results, since the addition of an $L_1$ penalty leads to a biased estimate
    of the true coefficients.
    (For example, if we rescale the variable $X_0$ by a factor $c > 1$, then the original weights are shrunk by a multiplicative factor of $1/ c$, 
    which affects the $\|W\|_1$ component of the loss in \eqref{eqn:full_f}.)
    To isolate the influence of the $L_1$ penalty, the results we present here (and in all the following sections) are for $\lambda=0$. We
    again stress that the qualitatively same behavior can be observed for $\lambda > 0$.

    \section{Further observations and remarks}

\begin{figure}
        \includegraphics[scale=0.37]{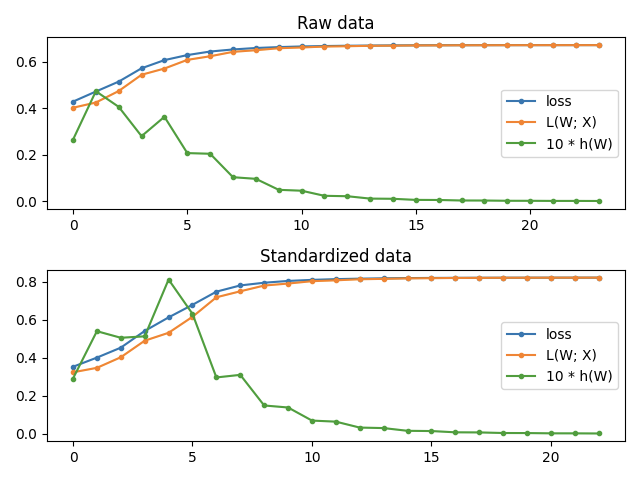}
        \caption{Plots of the loss terms.}
        \label{fig:loss_terms}
    \end{figure}

    Next, we analyze the dynamics of this iterative method to understand where things go wrong.
    It is convenient to visualize the results after each iteration of
    the inner update step. Figure~\ref{fig:loss_terms} shows the total loss
    (from Eq.~\eqref{eqn:opt_aug_lagrangian}) after each
    iteration, together with $\ell(W; X)$ and $h(W)$.
    After the first inner step (displayed at index 0), there is a small
    penalization for the lack of a DAG structure. In this case, the method
    ``overfits'' the data, leading to a fairly small loss $\ell(W; X)$, and hence
    a small overall loss in Eq.~\eqref{eqn:opt_aug_lagrangian}.
    As the loss contribution of $h(W)$ is
    increased with each step, the loss $\ell(W; X)$ tends to increase as well.
    This represents the tradeoff between fitting the data well and satisfying a DAG structure.
    Interestingly, also $h(W)$ has the tendency to increase initially. Once
    the multiplicative factors applied to $h(W)$ in
    Eq.~\eqref{eqn:opt_aug_lagrangian} increase, $h(W)$ becomes
    the predominant contributor to the loss at later iteration of the outer
    loop. This ensures that $h(W)$ eventually approaches 0.

\begin{figure}
        \includegraphics[trim=2cm 50 0 30, scale=0.27]{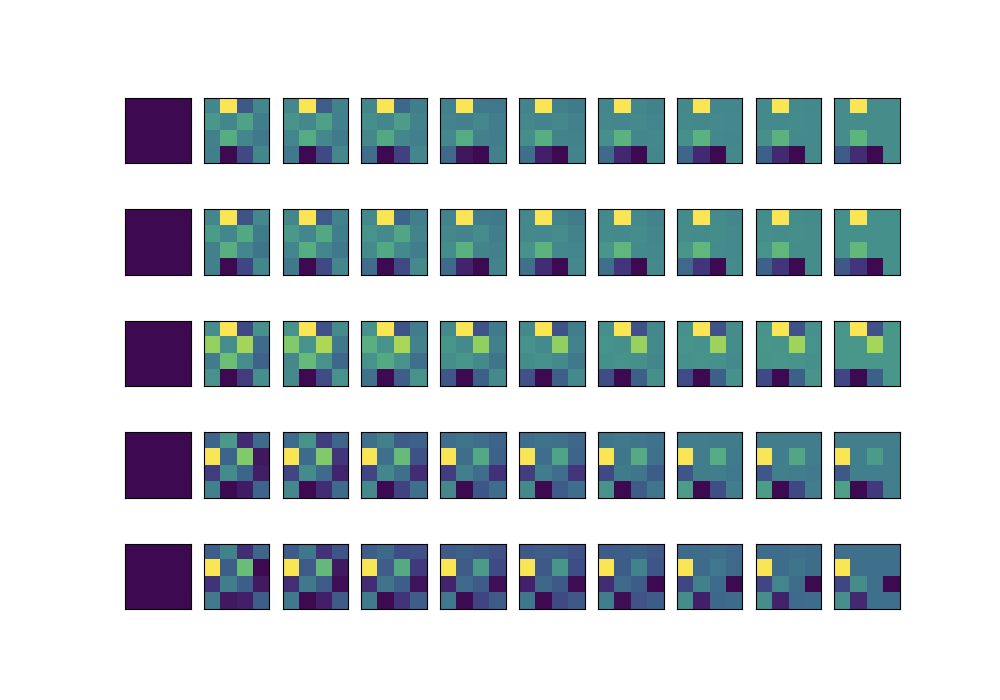}
        \caption{Each row represents the evolution of the weights with each inner
        update step, for different scaling factors (cf. main text).}
        \label{fig:multiple_histories}
    \end{figure}

    In Figure~\ref{fig:multiple_histories}, we visualize the dynamics of $W$ for different rescaling factors of the data.
    Each row represents the evolution of one set of weights.
    From top to bottom, the variables are incrementally rescaled - for the first row, we used the original data and in the last row we used the ``fully rescaled'' data,
    where each variable is normalized to unit variance.

    We observe that the first inner update step largely determines the
    final structure, with mainly smaller adjustment in the later steps.
    The first two rows lead to the correct result, but as we increase
    the rescaling factors for the variables, the method converges to a
    different (wrong) solution. Again, we can observe that for larger
    rescaling factors (close to unit variance for all variables), the
    results are flipped for certain variables.

    In summary, we observed the following points:
    The first inner update step largely determines the final structure.
    Moreover, this first update step seems to overfit the data and the result thereof is sensitive to the scaling of the variables.

    \section{Explanation}\label{sec:explanation}

    To better understand how the results depend on the scale,
    we consider the following toy example with two variables, where $X_0 =
    \gamma X_1$ for some $\gamma > 0$. Note that this is clearly
    a pathological example with no ``correct'' solution, since we can
    equivalently write $X_1 = \gamma^{-1} X_0$. Nonetheless, the example
    is useful for understanding the internal workings of the algorithm.

    In order to minimize $\ell(W; X)$, the method initially chooses positive
    weights for both off-diagonal entries of $W$ (in violation of the acyclicity
    constraint!). The strength of the violation of this constraint depends
    on the size of the penalty $h(W)$ compared to the remaining error in $\ell(W;
    X)$.

    Then, with each iteration, the weight of one of the two off-diagonal
    entries is pushed to zero, to compensate for the increasing
    penalty of $h(W)$ until one of the two terms is equal (or at least close) to 0. The
    crux here is that we aim to achieve the overall smallest loss of $F(W)$,
    which in this case is obtained by minimizing the sum of residual
    variances, which is the minimum of
    \begin{equation}\label{eqn:initial_loss_comp}
        \| X_0 \|^2 + \| X_1 - X_0 w_{01} \|^2 \quad \textrm{and} \quad \| X_1 \|^2 + \| X_0 - X_1 w_{10} \|^2,
    \end{equation}
    where $w_{01}$ and $w_{10}$ are the respective regression coefficients.
    For this example, the second term in both equations is zero, due to the perfect linear relationship $X_0 =
    \gamma X_1$.
    Therefore, the chosen direction solely depends on the variances of $X_0$ and $X_1$.
    When $\gamma > 1$, then $X_1$ has the
    larger variance, and thus the lower residual error is obtained for the
    choice $X_0 \rightarrow X_1$ and $X_1 \rightarrow X_0$ for
    $\gamma < 1$ (for $\gamma\approx 1$ the direction is subject to randomness). In other words, the variable with lower variance is
    identified as the driver of the variable with higher variance.

    For a general example, the weights after the initial update step will typically satisfy
    $|W_{ij}| > 0  \Leftrightarrow |W_{ji}| > 0$, with equal signs, whenever the variables are nonzero.
    This explains the initial ``overfitting'': The outcome of the first inner
    update returns a matrix that explains variables ``both ways''. In later iterations, once the
    penalty for $h(W)$ is increased, the method will force a selection of the weights to
    zero in order to obtain a valid DAG structure.

\begin{figure}
        \includegraphics[scale=0.45]{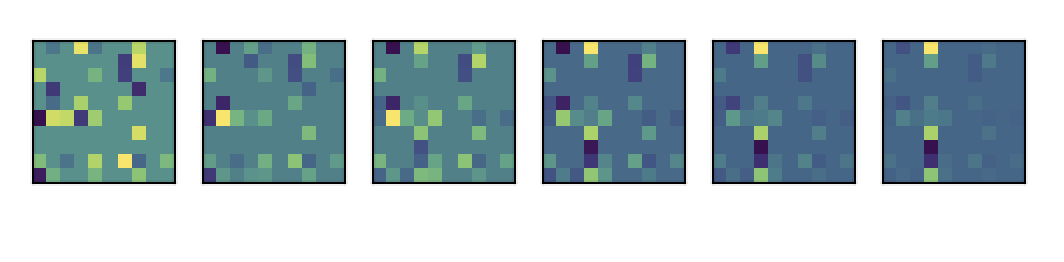}
        \caption{
            Multiple runs of the same example, where from left to right,
            the fourth variables is rescaled by the factors
            $[1, 2, 4, 8, 16, 32]$, respectively.}
        \label{fig:rescaling_one_var}
    \end{figure}

    Moreover, it is easy to show that increasing the variance for a
    single variable will lead to many edges pointing to this variable.
    This is because the $\|X - X W \|_F^2$ term in the loss function cannot be reduced to zero due to the acyclicity constraint.
    Some of the feature vectors $X_j$ will always be ``source variables'' for which all $X W$ terms are identically 0,
    whereas others will be ``sink variables'' for which the $X W$ terms are nonzero.
    When a feature vector $X_j$ has a significantly higher variance than other vectors, the NOTEARS method will prefer entries that
    reduce this variance as much as possible, thereby prioritizing inbound edges into the $j$-th node.
    This intuitively explains why the method lacks scale invariance, it is because nodes with high variance are preferred to be sinks as opposed to sources.
    As one performs a sweeping increase of the variance of a variable by changing its units, the resulting DAGs will change from having this variable
    act as a source to having this variable act as a sink.
    This is exemplified in Figure~\ref{fig:rescaling_one_var}, where the method is applied to the same data, with the exception that the variance for
    the fourth variable $X_3$ is doubled in each iteration.
    Iterating this argument, by rescaling all variables accordingly, one can then generate an arbitrary topological order
    (cf.~Appendix B in~\cite{peters2017elements}) for the identified DAG.
    
    The same issue persists for nonlinear extensions such as NOTEARS-MLP~\cite{zheng2020learning} (\url{https://github.com/xunzheng/notears}), which also depends on the Augmented Lagrangian approach described above.

    \section{Related work}\label{sec:relatedwork}
    
    Since we finished the initial version of this work, we became aware of the recent preprint \cite{reisach2021beware}, 
    which also addresses the point of scale dependence of continuous score methods from a more theoretical perspective. 
    This section of the paper links our findings with the literature and in particular reflects the discussion in \cite{reisach2021beware}. 

    Under suitable assumption on the data generation process combined with the structural form $X = XW + \epsilon$, the authors of \cite{loh2014high} 
    show that the global minimizer of the least square error in
    Equation \eqref{eqn:LS} does not always correspond to the true causal structure and 
    establish that Equation \eqref{eqn:LS} should be replaced by the Mahalanobis distance for the covariance of the noise terms $\Sigma := \operatorname{Cov}(\epsilon)$, i.e.
    \begin{equation}\label{eqn:ls_weighted}
        \ell(W; X, \Sigma) := \frac{1}{2n} \|(X - X W) \Sigma^{-1/2}\|_F^2
    \end{equation}
    to guarantee that the global minimizer of \eqref{eqn:ls_weighted} corresponds to the true causal structure.
    Since the true noise variance is not known for real world applications, one possibility is to estimate $\Sigma$ as the
    residual variance of $X - XW$. This is addressed by the authors of GOLEM~\cite{ng2020role}, which aim to improve
    on NOTEARS by replacing Equation \eqref{eqn:LS} by a Gaussian log-likelihood function. 
    They propose two versions with different plug-in estimators for the residual variance -- one with equal variances (GOLEM-EV)
    and one with non-equal variances (GOLEM-NV). An interesting point worth noting is that the likelihood contains the term $\log|\det(I - W)|$, which can be seen as a soft constraint for acyclicity and improves the performance of GOLEM over NOTEARS on synthetic data \cite{ng2020role}. The resulting likelihood functions are given by
    \begin{equation}\label{eqn:golem_llh_ev}
            \ell_{EV}(W; X) := \frac{d}{2}\log\bigl( \|X - X W\|_F^2\bigr) - \log|\det(I - W)|
    \end{equation}
    and 
    \begin{equation}\label{eqn:golem_llh_nv}
        \ell_{NV}(W; X) := \frac{1}{2}\sum_{i=0}^{d - 1}\log\bigl( \|(X - X W)_{i}\|^2\bigr) - \log|\det(I - W)|,
    \end{equation}
    where $(X - X W)_{i}$ is the $i$-th column of the residual matrix. 
    Note that we have $\ell_{NV}(W; X) \le \ell_{EV}(W; X) -\log(d)$ by an application of Jensen's inequality. Similar results to the ones in Section \ref{sec:explanation} apply for Equations \eqref{eqn:golem_llh_ev} and \eqref{eqn:golem_llh_nv} and in particular the same implication holds for Equation \eqref{eqn:initial_loss_comp}. 
    
    We refer the reader to \cite{reisach2021beware}, where the authors show that GOLEM-EV is -- just as NOTEARS --
    susceptible to rescaling of the data. GOLEM-NV is less affected by rescaling variables, but the overall performance of GOLEM-NV on unscaled data is worse.
    The authors stress that the performance of such methods on synthetic data can often be attributed to the way data are simulated
    for additive noise models (child nodes have a higher variance than their parents). They define 
    a baseline method, which identifies a DAG structure that corresponds to increasing variance and also propose
    a new measure for simulated data, which allows to identify to which extent child nodes in the simulated DAG have
    higher variances than their parent nodes.

    \section{Conclusion}

    We have seen that we can interpret NOTEARS as a method for finding parsimonious linear SEMs that best
    explain the data (as measured via the specified loss error).
    We showed that the results of the method heavily depend on the scale of the data.
    Similar to \cite{reisach2021beware}, we conclude that it is very important to carefully choose the scale of the data
    when one considers NOTEARS (or any method with similar loss function, 
    cf.~\cite{ng2020role,bhattacharya2021differentiable,vowels2021d}) for the identification of parsimonious models in real world scenarios.
    
    We here make the stronger claim that such methods are not suitable for identifying true causality from real world data.
    In deed, if the variables are on different scales (with different variances), then the edge orientation will be biased towards
    explaining variables with larger variances (subject to compatibility with the DAG-constraint).
    If the variables have (roughly) the same variance, then the resulting orientation of edges in the corresponding DAG will be
    subject to randomness.
    
    \begin{acknowledgements}
        We thank Hana Chockler and Andrew Lawrence for providing helpful comments on an early draft.
        We also thank Matthew Vowels, Alexander Reisach, Alexis Bellot and two anonymous referees for their careful reading of the manuscript and for many helpful comments and suggestions.
    \end{acknowledgements}

    \section*{Conflict of interest}
    The authors declare that they have no conflict of interest.

    \bibliographystyle{spmpsci}
    \bibliography{library}

\end{document}